%% file: main.tex
\documentclass[acmtog]{acmart}

\usepackage{booktabs} 

\usepackage[ruled]{algorithm2e} 

\SetAlFnt{\small}
\SetAlCapFnt{\small}
\SetAlCapNameFnt{\small}
\SetAlCapHSkip{0pt}

\acmJournal{TOG}

\begin{document}
\title{Personalized Speech-driven Expressive 3D Facial Animation Synthesis with Style Control}

\author{Elif Bozkurt}
\affiliation{%
  \institution{Huawei Turkey R\&D}
  \city{Istanbul}
  \country{Turkey}}
\email{elif.bozkurt@huawei.com}


\begin{abstract}
Different people have different facial expressions while speaking emotionally. A realistic facial animation system should consider such identity-specific speaking styles and facial idiosyncrasies to achieve high-degree of naturalness and plausibility. Existing approaches to personalized speech-driven 3D facial animation either use one-hot identity labels or rely-on person specific models which limit their scalability. We present a personalized speech-driven expressive 3D facial animation synthesis framework that models identity specific facial motion as latent representations (called as styles), and synthesizes novel animations given a speech input with the target style for various emotion categories. Our framework is trained in an end-to-end fashion and has a non-autoregressive encoder-decoder architecture with three main components: expression encoder, speech encoder and expression decoder. Since, expressive facial motion includes both identity-specific style and speech-related content information; expression encoder first disentangles facial motion sequences into style and content representations, respectively. Then, both of the speech encoder and the expression decoders input the extracted style information to update transformer layer weights during training phase. Our speech encoder also extracts speech phoneme label and duration information to achieve better synchrony within the non-autoregressive synthesis mechanism more effectively. Through detailed experiments, we demonstrate that our approach produces temporally coherent facial expressions from input speech while preserving the speaking styles of the target identities.

\end{abstract}

%
%

%
%


\maketitle

\input{sample2}

\end{document}

%% file: sample2.tex
\section{Introduction}
Synthesizing facial expressions considering identity relevant differences is crucial for realistic virtual avatars. Since humans express themselves differently for different emotion categories with smooth emotion transitions, users would prefer to engage more with virtual avatars those can portray such idiosyncrasies. Potential applications of such a system would take place in computer games, metaverse applications, animated movies and educational systems. 

Early methods for speech-driven facial expression synthesis based on deep-learning mainly consider training for a single identity \cite{karras2017,suwajanakorn2017,richard2021audio} and mostly require hours of data for learning the identity-specific facial motion sequences\cite{suwajanakorn2017,richard2021audio}. Although, the synthesized animation is very realistic, it is less practical for applications compared to methods that generalize over different identities\cite{richard2021meshtalk,thies2020nvp}. 

On the other hand, generalizing over different speakers would yield similar animations for different identities. Hence, the plausibility of such virtual avatars would decrease as users interact with them more often. One approach for modeling idiosyncrasies is using one-hot labels for each identity in the dataset\cite{VOCA2019,faceformer2022,Thambiraja2022Imitator}. Such approaches generally input the identity label along with the speech input and concatenate\cite{VOCA2019} or concatenate to latent representations in the generator\cite{faceformer2022,Thambiraja2022Imitator}. However, it is not guaranteed that the one-hot label representation information is carried out till the network output or facial motion peculiarities are fully-represented. Furthermore, using one-hot labels restricts the model adapting to unseen identities. Hence, few-shot or one-shot frameworks prefer to represent identities as a latent space\cite{abdal20233davatargan,chandran2022facial,erroll2022}. 

Intuitively, some frameworks extract identity and emotion information from the input speech sequence and synthesize expressive facial motion \cite{karras2017,Mittal_2020_WACV,peng2023emotalk}. However, speech to facial expression is a one-to-many mapping. Besides, it is challenging to learn all facial motion details only from speech input, considering not all facial parts (E.g. eye-brows) have full synchrony with speech as the orofacial area\cite{richard2021meshtalk}. Thus, some recent approaches extract the personalized motion information from the facial expression sequences and apply to the synthesis framework\cite{chai2022,aylagas2022voice2face}. Additionally, people often speak and express emotions simultaneously. So, expression disentanglement is required to achieve speech content independent facial expression representations or styles\cite{embed2021,chai2022}.

 A common approach for facial animation synthesis is using auto-regressive systems\cite{VOCA2019,faceformer2022,karras2017,richard2021meshtalk}. However, they may fail to model unseen emotion transitions based on the assumption that future expressions depend on previous ones. Thus, any change in emotion category should be learned specifically on the training dataset. On the other hand, non-autoregressive systems do not have aforementioned requirements \cite{Petrovich_2021_ICCV}. Moreover, they are unlikely accumulate errors over time and straightforward to parallelize incontrast to auto-regressive approaches. 
 
 Inspired by \cite{Petrovich_2021_ICCV}, we propose a non-autoregressive approach for synthesizing the facial expressions by considering duration and relative timing information in the decoder model. The proposed decoder model also has adaptive layers, which are beneficial for modulating the generated facial expressions with identity and emotion category specific styles globally. 

The main contributions of this work can be summarized as follows.
\begin{itemize}
\item We disentangle facial expressions into speech content dependent and identity-specific (style) components via adversarial classifiers
\item We estimate the phoneme labels and their durations from speech representations for better driving facial motion 
\item We model emotions and identity specific facial motion via style-aware speech encoders and expression decoders
\item We model emotion transitions via relative encoding and duration based modeling in the decoder
\end{itemize}

\section{Literature review}
\subsection{Speech-driven Facial Animation Synthesis}
Speech-driven 3D facial expression synthesis can be categorized into subject specific, generalized, and personalized based approaches, where the first only models a single person\cite{karras2017,suwajanakorn2017}, the second models various identities within a common representation \cite{richard2021meshtalk,aylagas2022}, and the latter learns idiosyncrasies for different identities \cite{bao2023learning,faceformer2022,VOCA2019,chai2022}. 
Cudeiro et al. \cite{VOCA2019} introduces the VOCA dataset, a 4D dataset with accompanying speech recordings and uses unnormalized  log probabilities of characters from DeepSpeech \cite{hannun2014deep} as input features in a system that animates 3D mesh vertices given a speaker identity. By conditioning the animations on different identities, their system can separate inter-speaker animation styles and generate personalized animations.

Richard et al. \cite{richard2021meshtalk} trains two separate encoders to learn
lip sync from audio and facial expressions respectively, fusing the
representations in a categorical latent vector that then animates a template
mesh. While the fidelity of the generated animations is impressive,
they train a separate model to generate audio-conditioned categorical
latent vectors since an expression sequence is not available at
inference time, thereby foregoing explicit control of speech style.

Most related to our work is  \cite{ma2023styletalk}, which first extracts phoneme sequences from speech to drive the video animation sequence, and extracts facial motion patterns of a style reference video to control style using adaptive style weights.  
Differently, our approach inputs style code in speech encoder and does not need phoneme labels during inference, also style is jointly modeled with phoneme duration information considering emotion transitions. Besides, we use adversarial identity classifier to disentangle facial motion styles. Moreover, our method is designed to produce expressive facial animations for a wide range of identities. Unlike \cite{VOCA2019, faceformer2022}, 
we aim to produce personalized animations with speaker-dependent
characteristics extracted from facial expressions rather than using one-hot identity/emotion representations. In contrast to \cite{richard2021meshtalk}, we maintain control of the generated animations by explicitly modeling expression style as a latent variable, optimized to
yield different emotion categories. In addition, by extracting phoneme and duration information from speech latent space, we avoid any information loss associated with discretizing to phonemic or visemic representation inputs. 

\subsection{Long sequences generation}
Synthesizing long and smooth sequences of animations is crucial for realism. However, naively increasing the maximum sequence length introduces increase in computational costs.  
\cite{tseng2023edge} proposes to enforce temporal consistency between multiple sequences such that they can be concatenated into a single longer sequence. However, such an approach has limitations for condition changes during motion synthesis, such as different dance style transitions, which should exist in the training set. On the other hand, recent motion in-betweening approaches consider motion infilling given previous and forthcoming motion frames \cite{motion_inbetween,oreshkin2022motion}. Additionally, text conditioned action motion synthesis approaches also consider generating long sequences by producing realistic human motions for a wide variety of actions and temporal compositions from language descriptions\cite{athanasiou2022teach}.   

 Our goal is to generate arbitrarily long sequences, such that each time window of the motion is potentially controlled with a different emotion category, and the transitions between windows are realistic and semantically matching the neighboring windows. Inspired from motion in-betweening literature \cite{motion_inbetween}, we model the transition from one emotion category to another by using masked-multihead attention and a learned relative position representation. 
 
\section{Proposed Method}
\begin{figure*}
  \includegraphics[width=\textwidth]{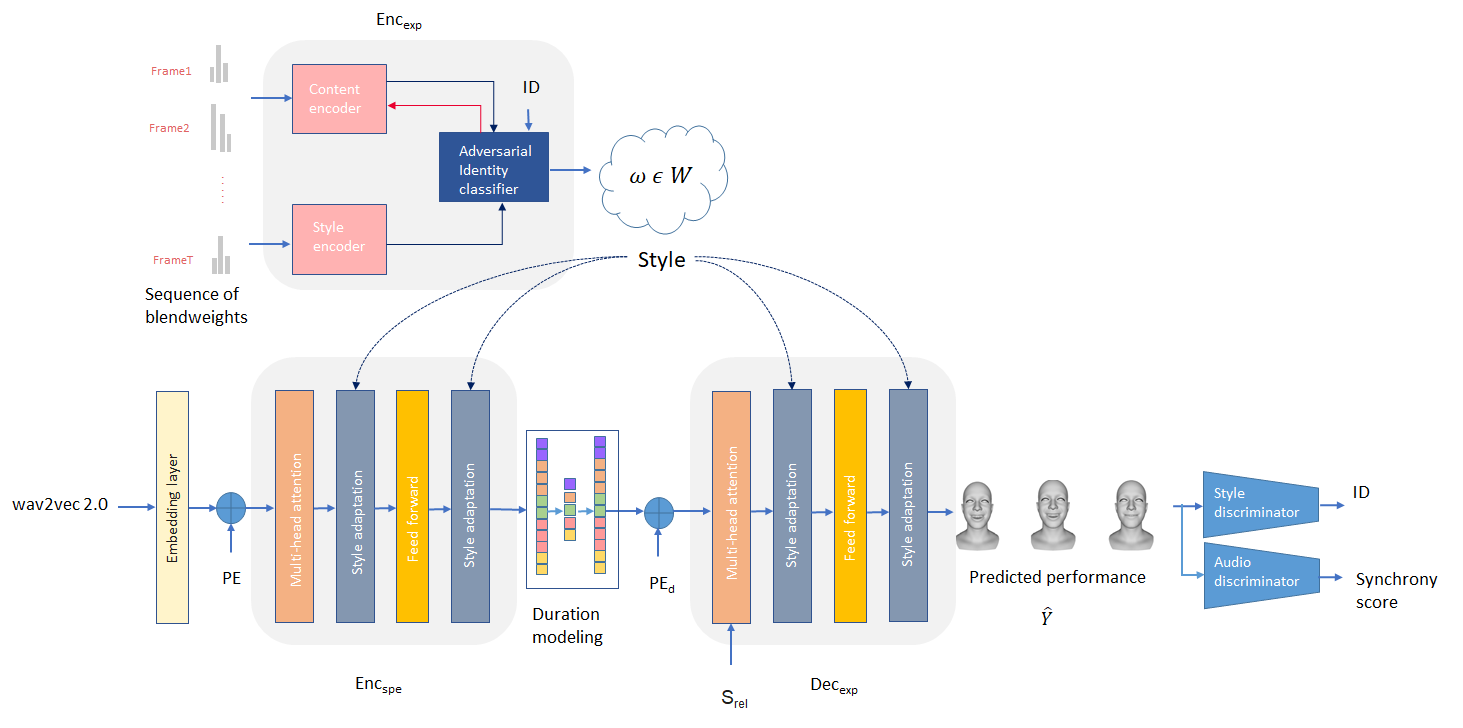}
  \caption{Our framework has three components: expression encoder, speech encoder and expression decoder. Expression encoder is composed of content and style encoders for expression disentanglement. We also use duration modeling to better align speech and expression sequences. Lastly, we use two discriminators to evaluate the synthesized sequences style and their synchrony with speech.}
  \label{fig:overview}
\end{figure*}

The goal of this work is to present a method that models a personalized, multi-modal distribution of facial expressions given a speech signal over time. Our model consists of three components, i.e., an expression encoder, a speech encoder and an expression decoder. Fig.~\ref{fig:overview} illustrates the an overview of our framework. During the training stage, all components of the model are conditioned on the same speaker and trained through the self-reconstruction loss of the expression sequence. At inference time, our model can achieve a personalized animation synthesis by injecting the target identity embedding both into the speech encoder and the expression decoder, respectively.  

\subsection{Expression Encoder}
We describe facial expressions as Facial Action Coding System (FACs) based blendshapes weighted by the corresponding blendweights. We use 41 blendshapes relevant to $\textit{eye-brow}$, $\textit{eye-lid}$, $\textit{lip}$, $\textit{chin}$, $\textit{cheek}$, and $\textit{jaw motion}$. Since blendweights also convey information related to speech content, we disentangle speech content related and identity related expression representations by using two identical encoders ${\textit{Enc}_{cont}}$ and ${\textit{Enc}_{sty}}$, where ${\textit{Enc}_{cont}}$ models content and ${\textit{Enc}_{sty}}$ models the style information.

Both of the encoders take a reference expression (blendweight) sequence $Y=[y_0, y_1, ..., y_T]$ of length $T$ frames as input. The goal of the style encoder ${\textit{Enc}_{sty}}$ is to extract a style vector $\textit{w} \in \mathbb{R}^N$ , which contains the style of given expression sequence $\textit{Y}$.  We use a 1D convolutional network with multi-head self attention to encode the global style information. Then, we use an identity classifier ${\textit{C}_{sty}}$, which classifies content and style encoder outputs into training speaker identities. 

To achieve correct expression disentanglement ${\textit{Enc}_{cont}}$ output should not contain any identity relevant representations.  Hence, we adopt an adversarial classification method that penalizes the content encoder for identity relevant outputs. Specifically, Gradient Reversal Layer (GRL) \cite{ganin2015unsupervised} is added before the identity classifier ${\textit{C}_{sty}}$, so that the gradient is reversed for ${\textit{Enc}_{cont}}$ during back-propagation.

To calculate the identity recognition accuracy at the output of the style encoder, we adopted the cross-entropy loss between ${\textit{Y}}$ and $\hat{\textit{Y}}$ as follows:

\begin{equation}
\label{eqn:01}
\mathcal{L}_{spk}=-\frac{1}{K}\sum_{i}^{K}{y_i log(C_{sty}(y_i))},
\end{equation}
where ${\textit{K}}$ is the number of speakers.

\subsection{Speech Encoder}
We use wav2vec 2.0 speech representations \cite{baevski2020wav2vec}  $X=[x_0, x_1, ..., x_T']$ of length $T'$ frames for better generalization to unseen speech data. Since, wav2vec 2.0 is a pretrained speech model which has been trained on a large-scale corpus \cite{librispeech} of unlabeled speech, it has rich learned phoneme information. Our speech encoder,  ${\textit{Enc}_{spe}}$ converts a sequence of wav2vec 2.0 embeddings and the style vector $w$ into hidden sequences ${\textit{H}_{w2v}}$ to better drive the synthesis. ${\textit{Enc}_{spe}}$ is composed of Feed-Forward Transformer blocks based on the Transformer architecture \cite{vaswani2017attention} with adaptation layers for personalization based on the style vector ${\textit{w}}$.

\begin{equation}
\label{eqn:enc_spe}
{H}_{w2v}={\textit{Enc}_{spe}({\textit{X}, {\textit{w}}})},
\end{equation}

Then, we use a duration model ${\textit{H}_{dur}}$ to regulate the length of the hidden phoneme sequence into the length of the expression sequences ${\textit{Y}}$. Finally, the expression decoder ${\textit{Dec}_{exp}}$ converts the length-regulated phoneme hidden sequence into predicted expression sequence ${\hat{\textit{Y}}}$.  

\subsection{Duration modeling}
\begin{figure}
  \includegraphics[width=80mm]{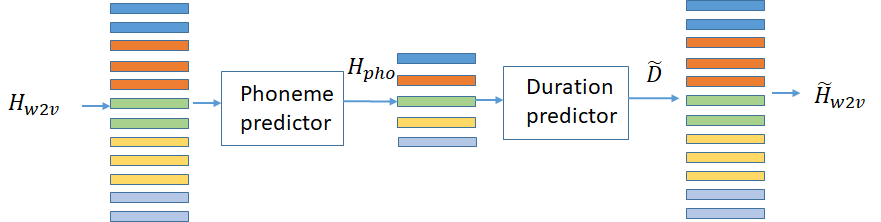}
  \caption{Duration modeling.}
  \label{fig:duration}
\end{figure}

In order to better align speech and facial expressions, we introduce a duration model for learning duration distributions for the phonemes as shown in \ref{fig:duration}. Since, vowels and consonants have different duration distributions, the conventional reconstruction would assign unit distribution for different phonemes. Moreover, since the introduced framework is non-autoregressive, learning duration distributions well would yield to more realistic motion sequences.

\noindent{\textbf{Phoneme-based downsampling}}
We assume the information for each phoneme duration does not vary much and use phoneme-based averaging to downsample the information.
We use the force alignment method for extracting duration information of the phonemes using \cite{mcauliffe2017montreal}.

Let ${\textit{H}}_{w2v}=[{\textit{a}}_{0},{\textit{a}}_{1}, ..., {\textit{a}}_{T'}]$ be the {\textit{wav2vec2.0}} hidden sequence  and 
${\textit{H}}_{pho}=[{\textit{h}}_{0},{\textit{h}}_{1}, ..., {\textit{h}}_{M}]$ be the phoneme hidden sequence, where ${\textit{T'}}$ and ${\textit{M}}$ are the lengths of the wav2vec2 and phoneme hidden sequences, respectively. In order to get phoneme duration based downsampling, we use a phoneme predictor ${\textit{P}_{pho}}$, which predicts the phoneme sequence corresponding to the ${\textit{H}_{w2v}}$. We use cross-entropy loss (\ref{eqn:01}) for the phoneme prediction with ground-truth phoneme labels ${\textit{P}}=[{\textit{p}_{0}}, {\textit{p}_{1}}, ..., {\textit{p}_{T'}}]$.

\begin{equation}
\label{eqn:pho}
\mathcal{L}_{pho}=-\frac{1}{T'}\sum_{i}^{T'}{p_i log(P_{pho}(h_i))},
\end{equation}
where ${\textit{T'}}$ is the length of the ${\textit{H}_{w2v}}$ hidden sequence. Then, we take average of the phoneme hidden sequence based on the corresponding duration values as,

\begin{equation}
\label{eqn:downsample}
{\widetilde{h}_{i}}=\frac{\sum_{t=l+1}^{l+d_i}{h_t}}{d_i}, for i=1,...,M
\end{equation}

\begin{equation}
\label{eqn:dur_mean} 
{l}=\sum_{j=0}^{i-1}{d_j} for {d}_{0}=0
\end{equation}

where ${\textit{M}}$ is the length of the wav2vec 2.0 hidden sequence ${\textit{H}_{w2v}}$.

Although we use ${\textit{P}}$ during the training stage, we use predicted phonemes and phoneme duration sequences during the inference time.

\noindent{\textbf{Phoneme duration based latent expansion}}
In order to upsample the phoneme hidden sequence back to the wav2vec hidden sequence, we use a duration predictor $P_d$ and a phoneme-level duration upsampler. We represent phoneme duration as the number of frames that belong to that particular phoneme class. 

Given the predicted duration sequence $\hat{D}$, the upsampler expands the hidden sequence by expanding based on $\hat{D}$. The upsampler predicts duration of each phoneme to regulate the length of hidden phoneme sequence into the length of speech frames. Fig.\ref{fig:duration} shows the downsampling and upsampling processes.

We use L1 loss with the ground-truth phoneme duration sequence ${\textit{D}}=[{\textit{d}_{0}}, {\textit{d}_{1}},...,{\textit{d}_{M}}]$ 

\begin{equation}
\label{eqn:04} 
\mathcal{L}_{d}=\frac{1}{M}\sum_{i=1}^{M}{|{P}_{d}(\widetilde{h}_{i})-{d}_{i}|}_{1} 
\end{equation}

We specify the predicted duration information in the position embedding, so that the expression decoder also inputs the duration information of the phoneme representations.

Input to the decoder is defined as 

\begin{equation}
\label{eqn:dec_input} 
H_{d}=\mathit{\hat{H}}_{w2v}+\mathit{PE}_{d} 
\end{equation}

where ${\textit{PE}_{d}}$ is the position embedding based on the duration values ${\hat{D}}$.

\subsection{Expression Decoder}
The expression decoder, ${\textit{Dec}_{exp}}$, aims to generate expression blendweights ${\hat{\textit{Y}}}$ given length-regulated phoneme hidden sequence ${\textit{H}_{d}}$ and a style vector ${\textit{w}}$, or exampler expression sequence. Inspired by the architecture of FastSpeech2 \cite{ren2022fastspeech}, which is a state of the art single-speaker non-autoregressive text-to-speech framework, we use the duration based temporal information in the transformer based decoder.

\begin{equation}
\label{eqn:dec_exp}
{H}_{exp}={\textit{Dec}_{exp}({\textit{H}_{d}, {\textit{w}}})},
\end{equation}

\noindent{\textbf{Style adaptation}}
Conventionally, the style vector is provided to the generator simply through either concatenation or the summation with the encoder output or the decoder input. On the contrary, we apply style through adaptive layers, where the adaptive layer receives the style vector, $w$, and predicts the gain and bias of the feature vector using linear layers. 

We apply layer normalization prior to the adaptation layer input and then, we compute the gain and bias with respect to the style vector, $w$. Gain and bias can adaptively perform scaling and shifting of the normalized input features based on the style vector. By utilizing adaptive layer, the generator can synthesize personalized expressions given the input speech and reference expression sequence as blendweights.

\begin{figure}
  \includegraphics[width=80mm]{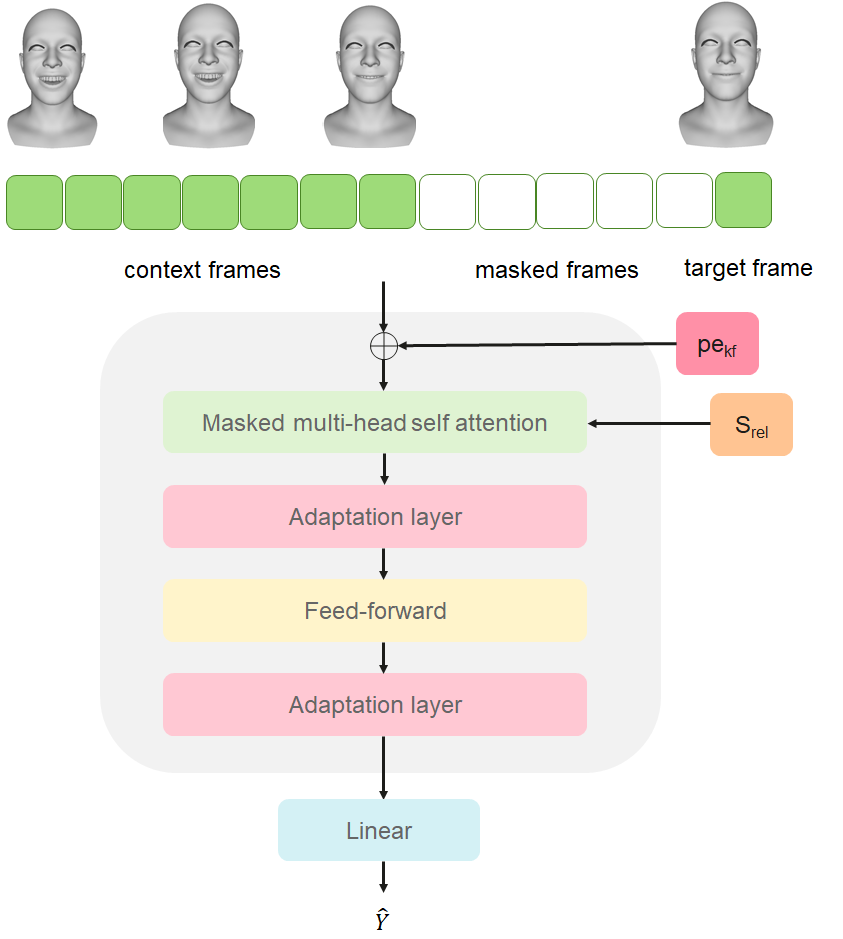}
  \caption{Expression decoder is based on Transformers. The figure shows a sample layer from the decoder. Context frames in the window are shown in green. The rightmost frame is the target frame. The masked frames are in white. Decoder learns to to fill the masked frames for emotion transition and inputs keyframe position embeddings ${pe}_{kf}$ at each layer and learned relative position embeddings ${S}_{rel}$ is added to self-attention computations.}
  \label{fig:decoder}
\end{figure}

\noindent{\textbf{Learned Relative Position Encoding}}
We use duration based position embedding at the beginning of the expression decoder. However, using only duration based position embedding may not be enough for emotion transitions when synthesizing a long sequence. Inspired from the recent motion in-betweening literature \cite{motion_inbetween}, we aim to model emotion transitions based on context frames and target frames as shown in Fig.\ref{fig:decoder} and use a learned relative position embedding to better model emotion transitions and apply it at every multi-head attention layer in the decoder. Similar to \cite{motion_inbetween}, we use the learned position embedding ${\textit{S}_{rel}}$ in the attention calculation as follows: 

\begin{equation}
\label{eqn:exp_att}
\mathit{Attention(Q,K,V)}=\mathit{softmax}\frac{{QK}^{T}+{S}_{rel}}{\sqrt{{d}_{K}}}V
\end{equation}

In other words, we define a context segment and a target frame as shown in the Fig.\ref{fig:decoder}. Then, we predict the in-between frames considering the style codes of the context window and the target frame. We assume target frame is the time step when there is an emotion transition.  We translate each element of the masked expression sequence into the number of frames remaining until the transition frame, which helps the generator to learn a natural transition between styles.

\subsection{Discriminators}
We use two discriminators, style discriminator and speech-expression synchrony discriminator, respectively. The style discriminator ${\textit{Dis}_{sty}}$, predicts whether the synthesized expression sequence follows the style of the target identity. The discriminator has a similar architecture with speech-encoder except it contains a set of prototypes ${\textit{S}={\textit{s}_{i}}}$, where $\textit{s}_{i} \in \mathbb{R}^N$  denotes the style prototype for the $i^{th}$ identity and $K$ is the number of identities in the training set. Given the style vector $w_{s}$ for style $s$ as input, the prototype $s_i$ is learned with the following classification loss 

\begin{equation}
\label{eqn:L_cls}
\mathcal{L}_{cls}=-\log\frac{exp({w}_{s}^{T}{s}_{i})}{{\sum}_{i'}{exp({w}_{s}^{T}{s}_{i'})}}
\end{equation}

The dot product between the style vector and all style prototypes is computed to produce style logits, followed by cross entropy loss that encourages the style prototype to represent the target identity's general style.

Moreover, to improve clustering of different styles in the latent space, we apply a classification at the discriminator output. The style discriminator then maps the generated expression sequence to a M-dimensional vector and compute a single scalar with the style prototype. The key idea here is to enforce the generated expression sequence to be gathered around the style prototype for each identity. In other words, the generator learns how to synthesize expressions that follows the general style of the target identity from a single short reference
expression sample. 

The second discriminator inputs the synthesized expression sequence and speech encoder output to distinguish synthesized expression from the original one given the speech representation as the condition. Discriminator evaluates the sequences at the frame-level and then uses the average of scores as the final score.

Both discriminators us LS-GAN based loss which is based on the least squares objective. The adversarial loss is computed as the sum of the two discriminator losses.

We use L1-based reconstruction and velocity losses:

\begin{equation}
\label{eqn:exp_rec}
\mathcal{L}_{rec}={\sum}_{i=1}^{T}{|{y}_{i}-\hat{y}_{i}|},
\end{equation}

\begin{equation}
\label{eqn:exp_vel}
{\mathcal{L}}_{vel}={\sum}_{i=1}^{T}{({y}_{i}-{y}_{i-1})-({\hat{y}}_{i}-{\hat{y}}_{i-1})}
\end{equation}

The total loss ${\mathcal{L}_{total}}$ is defined as:

\begin{equation}
\label{eqn:loss_total}
\mathcal{L}_{total}={\lambda}_{rec}\mathcal{L}_{rec} + {\lambda}_{vel}\mathcal{L}_{vel} + {\lambda}_{spk}\mathcal{L}_{spk} + {\lambda}_{dur}\mathcal{L}_{dur} + {\lambda}_{adv}\mathcal{L}_{adv}
\end{equation}

We set all $\lambda$ values as equal in the above equation.

\section{Experiments and Evaluation}
\subsection{Datasets and Implementations}
We require a naturalistic, multi-speaker, emotional and long-duration dataset to synthesize personalized and controllable animation sequences from speech. We train our framework on the BEAT dataset \cite{liu2022beat}. The BEAT dataset is a multimodal corpus of facial expressions, body motion, speech recordings, emotion annotations, perceptual annotations and text transcriptions. In the dataset, there are two scenarios for the data collection process: Conversational and improvisational data collection scenarios. We use the improvised recordings for training purposes and test on the conversational recordings, which are more spontaneous compared to the improvised ones.

We use four speakers from the BEAT dataset for training purposes and obtain four styles. We use five emotion categories in addition to neutral style for each speaker. The categories are $\textit{anger}$, $\textit{happiness}$, $\textit{sadness}$, $\textit{disgust}$, $\textit{fear}$, and $\textit{neutral}$. Content of the recordings depend on the emotion category and each speaker utters the same content for the same emotion category.

The frame rate for the expression sequences is 25 fps and the frame rate for the input speech wav2vec 2.0 features and phoneme sequences is 100 fps. We handle the frame rate difference in the duration modeling block so the input and output sequences are well-aligned. 

\noindent{\textbf{Implementation details}} We use Pytorch for implementing our framework. Our training method is end-to-end, where expression encoder, speech encoder, expression decoder and discriminators are trained synchronously. We also use a non-autoregressive training approach, where short duration windows of speech and expression sequences are used. The window length for expression sequence is 48 frames, and speech sequence is the corresponding window. Since our model predicts phoneme labels and their durations, the non-autoregressive approach is compatible to the auto-regressive approaches as mentioned in \cite{faceformer2022}. 

The expression encoder, which is composed of identical content and style encoders, aims to extract a global style from the input expression sequence. We use 1D convolutional layers and multi-head attention mechanism to extract the dynamic information specific to each identity in the dataset. Then, we use an identity classifier at the output of the style encoder that is composed of two linear layers with ReLU activation in between, and followed by a drop-out layer. The content encoder is penalized during back-propagation by using the GRL (gradient reversal) layer to ensure the expression content and style disentanglement is accurate.

The extracted style is injected to both the speech encoder and the expression decoder layers to have control on the output expression sequence's displayed style. Since we use style information in the speech encoder, as well, the emotional content of the speech input does not determine the output expression sequence's style, but only the injected style information does. Both speech encoder and the expression decoder has the same architecture with four transformer layers, with four-heads in the multi-head attention layers. However, we use different positional encoding methods for $\textit{Enc}_{spe}$ and $\textit{Dec}_{exp}$. We use the fixed sinusoidal position embedding for time representation in the  $\textit{Enc}_{spe}$. On the other hand, inspired from \cite{motion_inbetween} to better model relation between consecutive expression frames and to better model emotion transitions, we use relative position embedding in the expression decoder.

In a synthesized animation sequence, emotion transition from sadness to anger may be jerky, while from happiness to anger may be less jerky. Since we first use 1D CNNs to encode the speech input, the transitions between the synthesis window is smooth. The causal sequential module is composed of two causal 1D convolution layers with a LeakyReLU activation function for each. It serves to deal with longer temporal (by using windows) dependency and smooths the outputs from the audio input. We also use the same structure at the beginning of the expression decoder. It serves to deal with longer temporal dependency and smooth the outputs of the speech encoder. Since this structure considers extra history frames, temporal information longer than a speech window is covered. Additionally, a 1D convolution layer calculates a weighted sum of the inputs, so this convolutional architecture smooths inputs in nature. 

\subsection{Evaluation}
\begin{figure}
  \includegraphics[width=80mm]{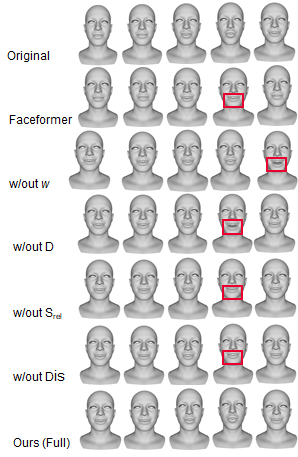}
  \caption{Comparison of synthesized sequences with ablation. Rectangles highlight errors on the lip shapes with respect to the original sequence shown at the top-most sequence.}
  \label{fig:comparison}
\end{figure}

Our framework is non-autoregressive and predicts phoneme duration, uses learned relative position encoding and adversarial learning to achieve realistic results. We compare our framework with the auto-regressive approaches proposed in \cite{VOCA2019} and \cite{faceformer2022} and evaluate by using the {\textit{lip vertex error}} method (LVE) introduced in \cite{richard2021meshtalk}, Frechet Inception Distance (${\textit{FID}}$), and audio-visual synchrony score introduced in (${\textit{Sync}}$) \cite{chung2017out} in Table \ref{tab:LVE}. LVE is the maximal $L2$ error of all lip vertices defined for each frame. We train VOCA \cite{VOCA2019} and Faceformer \cite{faceformer2022} on our dataset for accurate comparison. To understand benefit of each component in our framework, we also employ an ablation test. We select test speech recordings from the spontaneous recordings from the BEAT dataset. To visualize synthesized expression sequences we use the ICT-Facekit \cite{li2020learning} which has FACs compatible blendshapes as in the BEAT dataset. We report the average error over all testing sequences in Table\ref{tab:LVE} and demonstrate visual results in Fig.~\ref{fig:comparison}. 

The auto-regressive approach of \cite{faceformer2022} has an LVE value of $10.51$ on our test dataset. On the other hand, our full framework with style adaptation, phoneme duration estimation, learned relative position embedding and discriminators achieves the lowest LVE value as $9.12$. However, excluding style information during synthesis yields the highest LVE value as $10.8$. Using discriminators also helps to achieve lower LVE values since excluding them results in $10.65$ LVE value. Additionally, excluding duration model and learned relative position embeddings during synthesis yields increased LVE values on the test set as $9.84$ and $9.27$, respectively. We observe that learning identity specific motion sequences result in more accurate motion sequences.  

Similarly, our proposed framework improves realism of the sequences with an $FID$ score of $5.48$ and $Sync$ score of $10.40$. We observe again that using style information of different identities yields more accurate alignment of speech and facial motion compared to methods that generalize over all identities (w/out $w$).  

\begin{figure}
  \includegraphics[width=80mm]{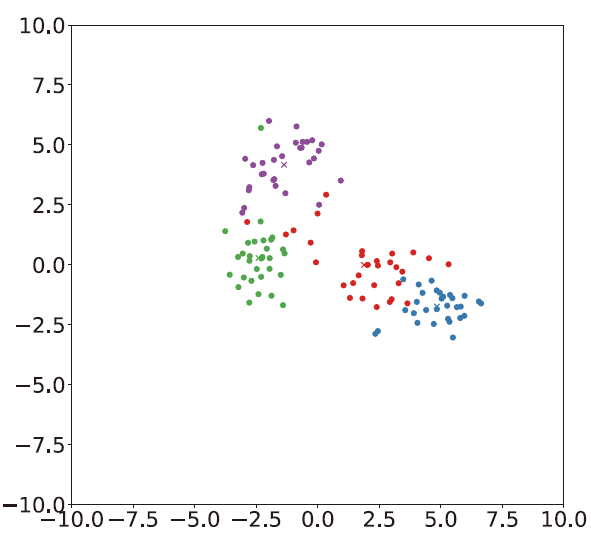}
  \caption{Styles of different identities in the training set. Each color represents an identity.}
  \label{fig:tsne}
\end{figure}

Furthermore, Fig.~\ref{fig:tsne} shows the distribution of different identities in the latent space. We observe that using style information of the identities $w$ based on disentanglement of facial expressions yields separate latent spaces.   

\begin{table}%
\caption{Objective comparisons with SOTA. Lip vertex error (LVE), Frechet Inception Distance (FID) and Sync (SyncNet) scores are presented.}
\label{tab:LVE}
\begin{minipage}{\columnwidth}
\begin{center}
\begin{tabular}{llll}
  \toprule
  Models   & LVE & FID & Sync\\ \midrule
  VOCA      & 12.28 & 7.28 & 9.22\\
  Faceformer     & 10.51 & 5.56 & 10.23\\
  Ours(w/out duration model)    & 9.84 & 6.29 & 10.12\\
  Ours(w/out style w)    & 10.8 & 6.96 & 10.18\\
  Ours(w/out relative pos.)    & 9.27 & 5.54 & 10.05\\
  Ours(w/out discriminator)   & 10.65 & 5.92 & 10.30\\
  Ours(Full)       & \textbf{9.12} & \textbf{5.48} & \textbf{10.40}\\
  \bottomrule
\end{tabular}
\end{center}
\bigskip\centering
\end{minipage}
\end{table}%

\section{Conclusions}
We introduced a personalized speech-driven expressive 3D facial animation synthesis system, which aims to model different facial expression sequences for different emotions considering the idiosyncrasies of the people. We additionally modeled emotion transitions between different categories. Our framework is non-autoregressive and predicts phoneme durations and uses learned relative position information to overcome any discontinuities in the synthesized sequences. We trained and tested our framework on the BEAT dataset and carried out ablation studies to observe importance of the each component in our framework. Comparisons with the state of the art auto-regressive \cite{VOCA2019} and \cite{faceformer2022} show that we can achieve better quality animations with our approach. Future work includes subjective tests on the generated animation sequences and more extensive experiments on animation duration control, insertion of motion sequences (i.e. laughter) for the selected time periods. Modeling the duration specific to phonemes would be useful when selecting specific expressions for the silence periods, changing duration of the animation.

%
%
%
%
\newpage
\bibliographystyle{ACM-Reference-Format}
